\documentclass[11pt,a4paper]{article}

\usepackage{authblk} 

\usepackage[hyperref]{acl2020}
\usepackage{times}
\usepackage{latexsym}

\usepackage{array,multirow,graphicx}
 \usepackage{float}
\usepackage{amssymb} 
\usepackage{graphicx} 
\usepackage{amsmath}  
\usepackage{mathtools}  
\usepackage{color}
\usepackage{arydshln}  
\usepackage{bbm} 
\usepackage{adjustbox}  

\usepackage{mathtools} 

\usepackage{microtype}

\usepackage{enumerate}
\usepackage{amsmath}
\usepackage[inline]{enumitem}
\usepackage[super]{nth}
\usepackage{booktabs}
\usepackage{textcomp}
\usepackage{multirow}
\usepackage{multicol}
\usepackage{tabularx}
\usepackage{amsfonts}
\usepackage{subfig}
\usepackage{float}

\aclfinalcopy 

\title{PARENTing via Model-Agnostic Reinforcement Learning to Correct Pathological Behaviors in Data-to-Text Generation}

\author[1,2]{Clément Rebuffel}
\author[1]{Laure Soulier}
\author[2]{Geoffrey Scoutheeten}
\author[1,3]{Patrick Gallinari}
\affil[1]{LIP6, Sorbonne Université, France}
\affil[2]{BNP Paribas, France}
\affil[3]{Criteo AI Lab, Paris}

\affil[ ]{\textit{firstname.lastname@\{lip6.fr,bnpparibas.com\}}}

\date{}

\begin{document}
\maketitle
\begin{abstract}

In language generation models conditioned by structured data, the classical training  via maximum likelihood almost always leads  models to pick up on dataset divergence (i.e., hallucinations or omissions), and to incorporate them erroneously in their own generations at inference. 
In this work, we build ontop of previous Reinforcement Learning based approaches and show that a model-agnostic framework relying on the recently introduced PARENT metric is efficient at reducing both hallucinations and omissions.
Evaluations on the widely used WikiBIO and WebNLG benchmarks demonstrate the effectiveness of this framework compared to state-of-the-art models.

\end{abstract}

\section{Introduction}

\begin{figure*}[t]
\centering
    \includegraphics[width=1\textwidth]{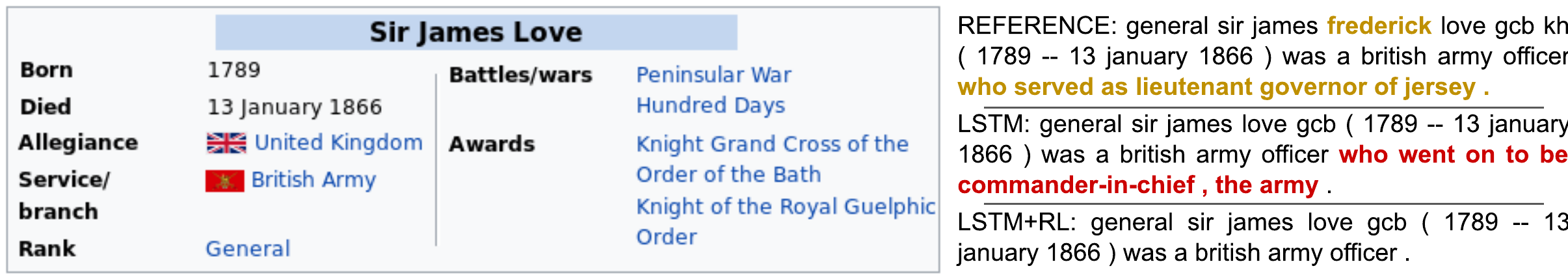}
    \caption{Example from WikiBIO where PARENTing (LSTML+RL) 
    led 
    to reduce hallucinated content. In yellow: divergent phrases in the reference; in red: hallucinated phrases from LSTM (without RL training) generation. 
    }
    \label{fig:jameslove}
\end{figure*}

Data-to-Text aims at generating natural language descriptions from structured data \citep{reiter2005}; fostered by recent advances on neural approaches  and 
the emergence of large scale datasets made of (structured-data, reference text) pairs \citep{lebret2016,gardent2017,wiseman2017}. Figure~\ref{fig:jameslove} illustrates an example from the WikiBIO dataset \citep{lebret2016}. These datasets are either hand-crafted via crowdworkers or automatically built by aligning sources found on the Internet. As such, 
reference texts might include divergences of two types, limiting the ability of generation models to produce realistic descriptions. First, reference texts might contain information not grounded in the source data;  especially for automatically constructed datasets, where references were not written with the source-data description task in mind. For instance, the phrase ``who served as lieutenant [...]'' in  Figure~\ref{fig:jameslove} has no basis in the associated infobox. Second, reference texts do not always cover the entirety of the table (items \textit{Battles/wars} in Figure \ref{fig:jameslove}). In most settings, this second point is referred to as content selection and is inherent of most data-to-text tasks. 
However, some hand-crafted datasets are designed where annotators are asked to transcribe every fields, with models also expected to do the same. In this case, incomplete references (\textit{i.e.} where some part of the source data is missing from the realization) can lead to models failing to learn to transcribe all information, and only partially cover data-sources at inference.

Divergence in training examples leads to hallucinated/omitted content in model output; which is a well-known problem in neural approaches for text generation \citep{rohrbach2018}. This problem arises both from the training procedure (training via maximum likelihood leads to language models strongly mimicking human behaviors), and from the testing protocols. Indeed, current standard metrics only measure similarity (such as BLEU \citep{papineni2002}, ROUGE \citep{lin2004}, METEOR \citep{banerjee2005}) to ground truth reference texts and do not fully capture relevance to the source data. 
Thus, there is no distinction between a mismatch caused by a paraphrase, poor lexicalization of content, or made-up/incorrect statement, leading to imperfect model selection.
While a number of work argue for the need for novel automatic evaluation method \citep{reiter2009,reiter2018,novikova2017}, to the best of our knowledge only \citet{wiseman2017} and \citet{dhingra2019} propose metrics based on both the reference and the source data. 

Recently, different regularization methods have also been proposed to mitigate the negative influence of divergences in reference texts. These approaches can be either at the dataset level \citep{dusek2019}, where authors propose techniques to clean/standardize instances; or at the training level \citep{tian2020}, where authors propose novel neural modules designed to limit hallucinations/omissions. However, these approaches are severely limited: e.g., they require significant annotation labor, model-specific tricks and/or manual tuning.  
Furthermore, virtually all proposed neural approaches still suffer from 1)~\textit{exposure bias} and 2)~\textit{inconsistency between train/test measurement}. Indeed, current neural models are trained via a mechanism called \textit{teacher forcing} \citep{williams1989}, where the decoder is fed the previous correct token, no matter its actual prediction~(1), in order to maximize the log-likelihood of the target sentence (including divergent phrases), but are evaluated through the previously discussed n-gram metrics~(2). See Section~\ref{subseq:parenting} for a more detailed discussion about this subject.\\


To the best of our knowledge, there have been few approaches \cite{liu2019,Liu2019b} focused on the training procedure. 
\citet{liu2019} train a hierarchical encoder-decoder on three auxiliary tasks (namely \textit{sequence labeling}, \textit{text auto-encoder} and \textit{multi-labeling classification}) which are meant to guide the decoding process. 
Closest to our work, \citet{Liu2019b} propose a novel neural module for constrained attention, along with a reinforcement learning (RL) training procedure based on BLEU and TFIDF. In our work, to remedy the above shortcomings and building upon the work of \citet{Liu2019b}, we show that no novel neural module is necessary to handle hallucinations and omissions. We propose a model-agnostic RL framework, called PARENTing, where pretrained models are further trained with a self-critical policy gradient algorithm \citep{rennie2016} to limit the impact of divergences in training examples on text generation. Specifically, we use the PARENT metric  \citep{dhingra2019} which exhibits a strong correlation with human evaluation, while being easier to use out of the box. We provide extensive automatic evaluations on two data-to-text model families (LSTMs and Transformers) on two widely used benchmarks (WikiBIO and WebNLG), as well as a more focused human evaluation 
on WikiBIO. 
We report new state of the art PARENT scores on both datasets while BLEU scores are on par with previous SOTA approaches, which shows that our framework efficiently reduces pathological behaviors while keeping generation fluent.

\section{Related Work}
\label{sec:related-work}
\subsection{Text Generation from Structured Data}

Data-to-text models 
can be classified in two broad categories: knowledge-based models and stochastic/data-driven approaches \citep{gatt2018}. 
The former approaches  \citep{reiter2000} are driven by experts' knowledge, leading to a pipeline architecture split into subtasks: content selection and text structuring (\textit{macroplanning}), sentence planning (\textit{microplanning}) and generating actual sentences (\textit{surface realisation}). While accurate and efficient at inference time, these methods require significant manual efforts for new use-cases.
In contrast, data-driven approaches tend to blur the distinction between these subtasks with end-to-end training on large corpora of aligned input data and output text \citep{gatt2018}. End-to-end methods have been proposed early, such as \cite{chen2008} who apply statistical machine translation techniques to the sportcasting domain. Recent neural approaches now propose to leverage progress in deep learning to represent these data into a semantic vector space (also called embedding space) and stem from the neural machine translation domain \citep{lebret2016,puduppully2019,wiseman2017}. \\ 
Particularly, \citet{wiseman2017} propose the now by default back-bone data-to-text architecture, with an attention mechanism \citep{bahdanau2014}, which computes a context focused on important elements from the input, and a copy mechanism \citep{gulcehre2016,see2017} to deal with unknown or rare words.

To address domain-specific constraints, a common approach is to build architectures that explicitly model the key-value structure of the input table~\citep{nie2018,liu2018,liu2019,Liu2019b,Rebuffel2020}. 
Additional work \citep{puduppully2019} introduces dynamic encoding updating, where the  model updates part of the source data encoding at each decoding step in order to 
accurately guide the decoder throughout generation. 

While these models produce fluent and domain-comprehensive outputs, several pathological behaviors have been identified, echoing similar issues in other text generation tasks (e.g. in image captioning \citep{rohrbach2018} or in summarization \citep{kryciski2019}).

\subsection{Pathological Hallucinations in Data-to-Text}

Training neural model on data-to-text tasks requires large corpora \citep{lebret2016,novikova2017,gardent2017,wiseman2017}. Different pathological behaviors arise from the  datasets, depending on the methodology underlying their construction. First, for hand-crafted datasets \citep{novikova2017,gardent2017}, crowdworkers sometimes fail to cover all information from the data source in reference text. Second, automatically constructed datasets from possibly different internet sources do not guarantee data sources and texts to be aligned completely. 
Both these limitations induce neural generation model to omit information in the first case or suffer from hallucinations (i.e., they mistakenly learn to generate ungrounded/false statements) in the other. 

To deal with these pathologies, previous work operate either at the dataset level, or at the training level. At the dataset level, \citet{dusek2019} show that cleaned data can significantly improve system ability to produce fact-accurate text. In a different direction, \citet{nie2019} apply a method similar to knowledge distillation \cite{hinton2015}: 
they train a Natural Language Understanding module to reconstruct tables from text references and show that a vanilla sequence-to-sequence model trained on the refined data has improved content correctness in both human and automatic evaluations. 
At the training level, 
\citet{wiseman2017}, for instance, propose to include a reconstruction loss aiming at reconstructing the source table from the hidden states of the decoder. 
In an other direction, \citet{perez-beltrachini2018} propose a classifying neural network, trained (using a manually annotated dataset) to label text tokens depending on their alignment with the associated table. They use these labels in an RL framework to generate sentences with a maximum of aligned tokens.
However these approaches are either costly in human labor or specific to hand-crafted datasets where the input data matches exactly the reference texts (thus deal with omissions but not hallucinations). Indeed reconstruction tasks are not compatible with the content selection subtask of Data-to-Text.

Proposing both a novel coverage-constrained attention and a BLEU/TFIDF-based reward, \citep{Liu2019b} constitutes a first approach to a model-agnostic framework. However their proposed coverage is still task specific (and goes against content-selection): while they increase the state-of-the-art BLEU on WikiBIO, they underperfom  encoder-decoder models on the PARENT benchmark.

Until recently, the NLG research community cruelly lacked ways to automatically evaluate model outputs.
Despite work on effective human evaluation \citep{amidei2019}, and on the need for better automated metrics \citep{reiter2009,novikova2017}, to the best of our knowledge, only \citet{wiseman2017} and \citet{dhingra2019} recently proposed improvement over the widely used BLEU. \citet{wiseman2017} propose to use an auxiliary neural model, trained to extract structured records from the generated text for evaluation. Two texts can then be compared through their sequences of extracted records. This information retrieval-based approach suffers from domain specificity, as the released model only works in the closed-domain of basketball journalistic summaries, and requires precise tagging of gold references which can be impossible to provide in most settings. Furthermore, \citet{dhingra2019} propose a new metric PARENT, and show that this metric strongly correlates with human annotators and can replace previous n-gram- and information retrieval-based metrics.

Our contribution differs from previous work in several aspects. First, our proposed framework is model-agnostic and can be used with any neural model. Second, instead of focusing on only one domain and/or one issue (e.g., omissions in hand-crafted datasets or hallucinations in automatically constructed datasets), it is setting agnostic and tackles both hallucinations and omissions at once by leveraging the PARENT F-score (which combines precision and coverage against the source data). Finally, no manual preprocessing or pre-tagging is required: models are trained via a flexible training protocol and distantiate themselves from faulty training examples.

\section{Model-Agnostic Reinforcement Learning for Reducing Divergences}
\label{sec:model}
We propose PARENTing, a model-agnostic RL framework for data-to-text aiming at reducing divergences. It is based on the  self-critical policy gradient algorithm \citep{paulus2017} and leverages the PARENT metric~\cite{dhingra2019}. 

\subsection{Background and Notations}

\paragraph{Notations.} 
We consider the general setting of data-to-text and the notations introduced by \citet{dhingra2019}. Let us consider a dataset of $J$ pairs (structured-data, reference), denoted $\mathcal{D} = \{(T_j, R_j)\}_{j=1}^J$ where:

\hspace{0.5cm} $\bullet$ $T \coloneqq \{ r_k \}_{k=1}^K$ is a collection of $K$ records (\textit{entity}, \textit{attribute}, \textit{value}), where $K$ is variable among instances; 

\hspace{0.5cm} $\bullet$  $R \coloneqq [y_1^*, ..., y_L^*]$ is the reference text associated to $T$, composed of L tokens $y^*$, where $L$ is variable among instances; 

We also consider a data-to-text neural model, denoted $f_\theta$ (where $\theta$ are the model parameters), pretrained to maximize the likelihood of the reference, via teacher forcing \citep{williams1989}: 
\begin{equation}
    \mathcal{L}_{ml} = - \sum_{t=1}^L \log f(y_t^* ~|~ y_{t-1}^*, ..., y_1^*, T, \theta)
    \label{eq:cross-entropy-loss}
\end{equation}


\paragraph{PARENT metric.}
 PARENT (\textbf{P}recision \textbf{A}nd \textbf{R}ecall of \textbf{E}ntailed \textbf{N}-grams from the \textbf{T}able) \cite{dhingra2019}  aims at evaluating the precision and recall/coverage of a candidate generation $G$ given the (source table, reference) pair $(T, R)$ via n-grams ($n=1,..., 4$) comparison. This metric is divided in three scores:
 
\hspace{0.5cm} $\bullet$  Entailed precision $\mathbf{E}_p$ is the fraction of n-grams from $G$ which are either found in $R$ or $T$;

\hspace{0.5cm} $\bullet$  Entailed recall/coverage $\mathbf{E}_r$. Recall $E_r(R)$ is the fraction of n-grams from $R \cap T$ which are found in $G$; Coverage $E_r(T)$ is the fraction of n-grams from $T$ which are found in $G$. Recall and coverage are combined using a geometric average: 
    \begin{equation}
        \mathbf{E}_r \coloneqq E_r(R)^\lambda E_r(T)^{1-\lambda}
        \label{eq:parent-recall-lambda}
    \end{equation}
\hspace{0.5cm} $\bullet$  F-score:  combination of precision and recall.

\subsection{Overview and Research Objectives}
Our framework for reducing pathological behaviors is based on the following research objectives:

\hspace{0.5cm} $\bullet$   O1: the framework should be generic and should work with any neural model; 

\hspace{0.5cm} $\bullet$   O2: the model should try and distantiate itself from the reference enough to stop mimicking problematic behaviors; 

\hspace{0.5cm} $\bullet$   O3: by combining precision, recall and coverage, PARENT is a good proxy for human assessment of a candidate text against its source data and reference \cite{dhingra2019,tian2020}; 

\hspace{0.5cm} $\bullet$   O4: discrete metrics can be gamed to artificially increase the score while not gaining in readability or relevance \cite{liu2016}.

Therefore, we propose a training protocol similar to \cite{paulus2017}, with a mixed objective function combining the standard maximum-likelihood loss $\mathcal{L}_{ml}$ with a custom reinforcement loss $\mathcal{L}_{rl}$. This ensures that models do not lose fluency by gaming the discrete metric (objective O4):
\begin{equation}
    \mathcal{L} \coloneqq \gamma \mathcal{L}_{rl} + (1 - \gamma)\mathcal{L}_{ml}
    \label{eq:mixed-objective}
\end{equation}
where $\gamma$ is a weight factor.
We note that O1 is satisfied, as this loss function $\mathcal{L}_{rl}$ can be applied to train any neural model $f_\theta$. In what follows, we give a description of the proposed reinforcement learning framework.

\subsection{PARENTing: Self-critical Gradient Policy Learning}
\label{subseq:parenting}

Numerous work \citep{wu2016,rennie2016} have outlined that training via teacher-forcing (maximizing the log-likelihood of reference texts) does not always produce the best results on evaluation metrics. 
This is in part due to \textit{exposure bias} \citep{ranzato1015}, where models are trained using the true gold sequence during training and are  never exposed to their possible mistakes. 

We therefore propose to alter the standard rigid training protocol and further train models via reinforcement learning as a counter-measure to these issues, where models can now learn a more flexible policy based on a metric more representative of human judgment, satisfying objective O2. Following objective O3, we shape our reward around the PARENT metric which has been shown to strongly correlate with human judgement in term of precision and recall of a generated text against a source table and a reference. Models are somewhat over-fitted to our training set due to pretraining, and are hence at risk of earning high rewards on easy examples (i.e. with faithfull reference targets) and poor rewards on hard examples (i.e. with divergent reference targets). To deal with this issue and ensure that the reward reflects the actual improvement made over the pretraining, we propose to follow a growing body of work in text summarization \citep{paulus2017,scialom2019} and apply the self-critical policy gradient training protocol  \citep{rennie2016}, using the REINFORCE \cite{williams1991} algorithm. 

More particularly, models are now sampled using their Markov property (that is one token at a time, and computing the next distribution given the previous chosen token). A first candidate sequence $Y^c$ is randomly sampled following the outputed distribution. A second baseline sequence $Y^b$ is generated, this time via greedy decoding (mimicking beam search generation during inference, with a beam of size $1$). This baseline sequence acts has a difficulty proxy of the current training instance. The reward given to the candidate sequence is the improvement in PARENT score it brings over the baseline sequence: 
\begin{equation}
    r(Y^c) = \text{PARENT}(Y^c) - \text{PARENT}(Y^b)
    \label{eq:reward-shaping}
\end{equation}

Finally, the loss to be minimized during this part of training is:
\begin{equation}
    \mathcal{L}_{rl} = - r(Y^c) \sum_{t=1}^L \log f(y_t^c ~|~ y_{t-1}^c, ..., y_1^c, T, \theta)
    \label{eq:rl-loss}
\end{equation}

Minimizing Equation~\ref{eq:rl-loss} leads to increase reward expectation. Indeed, we maximize the conditional likelihood of the candidate sequence $Y^c$ when it obtains a higher reward than the baseline sequence $Y^b$, or on the contrary we decrease its likelihood in case of a lower reward.

\section{Experimental setup}
\label{sec:experimental-setup}

\subsection{Data-to-text benchmarks}


\paragraph{WikiBIO} \citep{lebret2016} This dataset contains $728,321$ infoboxes, automatically paired with the first sentence of the corresponding article of the English Wikipedia. we follow the data partition introduced with the dataset which yields $80\%$ of all instances for the training set, $10\%$ for the development set and $10\%$ for the evaluation set. Reference texts are of average length $26$ words while infoboxes have on average 12 non-empty fields. This dataset has been built automatically from sources that were not meant for a text-generation task and contains a significative amount of divergence between the source data and the target descriptions ($62\%$ of the references mention extra information not grounded in the  infobox \cite{dhingra2019}).

\paragraph{WebNLG} \cite{gardent2017} This dataset contains $35,970$ sets of RDF records mapped to natural language descriptions. Each set has up to $7$ records, and one or more gold references of average size $22$ words. We follow the partition introduced with the dataset, which yields $1612$/$1619$ instances as a development/evaluation set. This dataset has been hand-crafted specifically for the task of surface realization and systems are expected to summarize all records. Note that here we compare ourselves on the \textit{seen} partition, where every attribute  is been seen during training (however, entities and values can be new).

\subsection{Evaluation metrics}

We evaluate our approach using both automated metrics and human judgment. We report BLEU scores \citep{papineni2002} as well as PARENT (precision, recall and F1) scores \citep{dhingra2019}. For all scores higher is better. While BLEU is the historical metric in all text generation tasks, PARENT scores have a significantly stronger correlation with human evaluators \cite{dhingra2019} ($0.478$ vs.  $0.913$ for BLEU and PARENT resp.).

We perform qualitative evaluation following the best practices outlined by \citep{Lee2019}. Our human annotators are males and females from several countries across Europe, between 20 and 55 years old and proficient in English. 
Annotators are shown a randomly selected table, together with the corresponding descriptions, both from the dataset and the models that are being evaluated. Annotators are asked, for each sentence, to score its fluency (as \textit{Fluent}, \textit{Mostly fluent}, or \textit{Not fluent}) factualness (likewise), and coverage (in terms of the number of realized rows). Sentences are shuffled to avoid any bias. 
Following \citet{tian2020}, we first tasked three expert annotators to annotate a pilot batch of 50 sentences. Once assured all Inter-Annotator Agreements were approx. $78$\%, 
we asked several annotators to annotate an additional sentence sample to reach 100 instances (where each instance consists of one table and three associated outputs). 

\begin{table*}[t]
\centering
\begin{adjustbox}{max width=0.9\textwidth}
\begin{tabular}{p{0.02cm}l|cccc@{}|cccc@{}}
\hline
&\multicolumn{1}{c|}{Model} & \multicolumn{4}{c|}{WikiBIO}                                               & \multicolumn{4}{c}{WebNLG}                                                \\ 
&                           & \multicolumn{1}{l}{BLEU} & \multicolumn{3}{l|}{PARENT (Prec. / Rec. / F1)} & \multicolumn{1}{l}{BLEU} & \multicolumn{3}{l}{PARENT (Prec. / Rec. / F1)} \\
                           \hline
\parbox[t]{2mm}{\multirow{4}{*}{\rotatebox[origin=c]{90}{Baselines}}}
&S2S+FA+RL & \textbf{45.49}                    & 76.1 & 45.9 & 54.8         & -                        & -              & -             & -             \\
&Confident PG               & 38.10                    & 79.52          & 40.60          & 51.38         & -                        & -              & -             & -             \\
&GCN                        & -                        & -              & -              & -             & 55.9                     & -              & -             & -             \\ 
&Gardent-LSTM & - &- &- & - & 54.03 &- &- &- \\\hline
\parbox[t]{2mm}{\multirow{4}{*}{\rotatebox[origin=c]{90}{Our models}}}&LSTM                & 42.80                    & 78.70          & 45.16          & 55.10         & 54.9                     & 73.15          & 69.91         & 69.67         \\
&LSTM+RL             & 44.17*                 & 80.01*       & \textbf{46.60}*       & \textbf{56.72}*      & \textbf{63.20}*                 & 74.51         & \textbf{70.68}*        & \textbf{71.27}*        \\
&Transformer &      41.09                    &    80.02            &     44.31           &   54.74            & 53.45                    & \textbf{75.17}          & 64.15         & 67.38         \\
&Transformer+RL &       42.40                   &    \textbf{80.37}            &     45.83*           &    56.15*           & 57.07*                  & 75.06          & 66.70*        & 69.01*\\
\hline 
\end{tabular}
\end{adjustbox}
\caption{Evaluation on WikiBIO and WebNLG. *: p-value$< 0.001$ of Student T-test  comparing  with/without RL.}
\label{table:results}
\end{table*}

\subsection{Scenarios and Baselines}

We measure the impact of our framework on two families of models: 

$\bullet$ \textit{LSTMs.} Our implementation of \citep{see2017}. It is the back-bone data-to-text model based on a bi-LSTM with attention mechanism and augmented with a conditional copy mechanism to deal with rare or unseen words. 

$\bullet$ \textit{Transformers.} Our implementation of \cite{vaswani2017}, the  transformer encoder-decoder, augmented with a conditional copy mechanism. 

These models are denoted \textbf{LSTM} or \textbf{Transformer}  when trained via maximum likelihood and \textbf{LSTM+RL} or \textbf{Transformer+RL} when further trained using the  PARENTing framework.
\vspace{1mm}

We also report  SOTA models for each dataset respectively (i.e., achieving the strongest score either BLEU or PARENT) : 

$\bullet$ For WikiBIO, we report the BLEU and PARENT scores of two baselines: 
1)~\textbf{S2S+FA+RL} \cite{Liu2019b} which uses a standard encoder-decoder structure, with an attention mechanism constrained to cover all table attributes and an RL training procdure with a reward shaped by BLEU and TFIDF;
2)~\textbf{Confident PG} 
\cite{tian2020}: a  neural module which assigns a confidence score to each output words, and trims the generated sequence from any word below a specified threshold. They report higher precision 
but lower fluency.

$\bullet$  For WebNLG, we report the BLEU score of \textbf{GCN}  \citet{marcheggiani2018}. They propose a graph convolutional network which explicitly models the structure of graph-like data. For additional context, we also report a baseline score introduced by the original paper  \citet{gardent2017}. The used model \textbf{Gardent-LSTM} is the same as our scenario LSTM.







\subsection{Implementations details}

We describe here key implementation details (other details needed for reproducibility will be given alongside the code if accepted). 
We set the $\lambda$ of Equation~(\ref{eq:parent-recall-lambda}) to $1$ during training. This was done 1)~because coverage is against the content selection task on most data-to-text tasks 2)~to reduce the computing cost, as coverage is obtained by computing Longest Common Subsequence for all n-grams contained in the table. We note however that we kept $\lambda = 0.5$ for evaluation following \citet{dhingra2019,tian2020}.
Preliminary experiment on $\gamma$ from Equation~\ref{eq:mixed-objective} showed that the initial value of $0.9987$ proposed by \citet{paulus2017} 
was not satisfying: fluency dropped drastically and while models obtained significantly higher PARENT scores, BLEU score was at less than half what previous models were able to achieve. We therefore used a more conservative value of $\gamma = 0.9$.
Inputs were fed to the neural networks following \citet{lebret2016}: each word is represented as a 4-tuple (value, field, p+, p-) where p+ (resp. p-) is the position (resp. reverse position) of value in field. For example, the line (Name, Barack Obama) is presented as [(Name, Barack 1, 2), (Name, Obama, 2, 1)]. In WebNLG, where tables include several entities, a $5^{th}$ element was introduced for entity index, as well as tokens for the entities' names.
Models are first trained via maximum likelihood training. We select the best performing checkpoint given a development set and start the mixed-objective training from there.
We implemented our framework using OpenNMT \cite{opennmt}. 
Data and code are available online: \href{https://github.com/KaijuML/PARENTing-rl}{https://github.com/KaijuML/PARENTing-rl}



\section{Results}
\label{sec:results}

Table~\ref{table:results} summarizes the BLEU and PARENT scores obtained by the baselines and our scenarios on  WikiBIO and WebNLG benchmarks. 
Please note that while no previous work report PARENT scores on WebNLG, our scenario \textit{LSTM} is a re-implementation of the baseline \textit{Gardent-LSTM}. The obtained and reported BLEU scores being very close, we can consider that our PARENT scores are also the ones of \textit{Gardent-LSTM}.

From a general point of view, We can see that our PARENTed models \textit{LSTM+RL} and \textit{Transformer+RL} obtain generally higher BLEU and PARENT metrics over all scenarios and baselines --- except the BLEU score for WikiBIO. 
More particularly, we can outline the following statements.

$\bullet$ The comparison of our scenarios (without/with our PARENTing framework) outlines increases in score ranging from $+1.6\%$ to $+3.4\%$ on WikiBIO, and from $+1.1\%$ to $+15\%$ on WebNLG; with significant improvements in $12/16$ comparison settings. This suggests that PARENTed models learn to describe source data with more precision (reduced hallucinations) and with greater details (increased recall/coverage), as can be seen in Figure~\ref{fig:jameslove}.

$\bullet$ BLEU scores are on par with baselines on WikiBIO, and significantly better than the strongest model on WebNLG.  
Despite starting close to the baseline in terms of BLEU, PARENTing our models leads to new state of the art BLEU of $63.20$, compared to a previous $55.9$ for GCN, representing a $13\%$ relative increase. This shows that models learn to lexicalize content more adequately than through maximum-likelihood training.


$\bullet$ More importantly, PARENTed models overperfom all baselines on PARENT scores: on WikiBIO, our reinforced \textit{LSTM} model achieves state of the art performance on F-score, increasing the previous score by $7\%$ (from $52.81$ to $56.72$). PARENTed models also achieve better precision than both \textit{S2S+FA+RL} and \textit{Confident-PG}: $80.01/80.37$ (for \textit{LSTM+RL} and \textit{Transformer+RL} resp.) against respectively $76.1$ and $79.52$. Contrasting with \textit{Confident-PG} which sacrifices fluency for faithfulness, our models referring more precisely to information from the table (\textit{i.e.} reducing hallucinations) does not come at the cost of less fluent output, as reflected by the BLEU scores of $44.17/42.40$ against $38.1$ for \textit{Confident-PG}.
We  note that beyond precision, PARENTed models are trained simultaneously to retrieve more relevant information (\textit{i.e.} reducing omissions), as they  achieve SotA recall performance ($46.60/45.83$), again contrasting with \textit{Confident-PG} which loses more than $11\%$ to \textit{S2S+FA+RL}, with $40.6$ against $44.02$.

$\bullet$ Altogether,  the previous statements assess  the model-agnosticity of our framework, as both model family (\textit{LSTM} and \textit{Transformer}-based scenarios) showed improvements on both datasets when fine-tuned with our PARENTing framework.


$\bullet$ We observe that our pre-trained scenarios (\textit{LSTM} and \textit{Transformer}) generally obtain higher results on WebNLG than on WikiBIO. This is due to the nature of datasets: WebNLG is hand-crafted with the explicit goal of full transcription of tables while WikiBIO is build automatically without rigorous alignment of data sources and reference texts. Despite some inevitable divergences, WebNLG is thus less noisy than WikiBIO.

\paragraph{Qualitative Evaluations.}
\label{subsec:qualitative-eval}
    

\begin{table}[t]
    \begin{tabular}{lc}
    \hline
     & LSTM scenario \\
    \hline
    Avg length without RL  &$19.17$\\
    Avg length with RL & $19.78$\\
    Avg  length $\Delta$&  $0.61$\\
    $\Delta$ of F-measure scores & $1.62$ \\
    Correlation &  0.21* ($p < 0.001$)\\ 
    \hline
    \end{tabular}
    \caption{Comparative statistics between \textit{LSTM} and \textit{LSTM+RL}  on WikiBIO. $\Delta$ for variation scores.}
    \label{table:deltas}
\end{table}


We aim to provide insight into what our PARENTing framework brings to models. Specifically, one might assume that a model could trivially learn to shorten output in order to increase precision, or on the contrary, to increase generation length to easily increase coverage by mechanically quoting tables more. We therefore 1)~check for the framework impact on global generation length; 2)~provide a more detail analysis on length distribution vs. effectiveness.


In this section, we focus exclusively on WikiBIO as it is the most challenging setting (larger vocabulary, more noisy, and content selection needed to generate biographies). 
To make results more readable, we focus on  (\textit{LSTM}, \textit{LSTM+RL}) models.

We first report in Table \ref{table:deltas} comparative statistics of generation length and score variations.
We first note that on WikiBIO, there is no significant changes in sentence lengths after RL training ($19.17$ vs $19.78$).
We further find a correlation of $0.2$ between the variation in length and the one in PARENT F-score which does not allow to conclude that longer texts lead to better scores. This suggests that the RL-trained model seems to have better performance even when generating shorter texts. 

We investigate further the impact of PARENTing on length distribution, and its influence on hallucinations/omissions (respectively measured by precision/recall).
To do so, generated texts are splitted in two broad categories, short and long, using a KMeans algorithm calibrated on the length of human references. We exhibit two clusters, texts below/above $30$ words, and compute PARENT scores conditioned on these clusters (see Table~\ref{table:parent-vs-length}).

\begin{table}[t]
    \begin{tabular}{lcc|cc}
              & \multicolumn{2}{c|}{LSTM}  & \multicolumn{2}{c}{LSTM+RL} \\ \cline{2-5} 
              & Short       & Long        & Short        & Long        \\ \hline
    Precision & 78.75       & 77.76*      & 80.04        & 80.03       \\
    Recall    & 44.82       & 52.01*      & 46.27        & 52.41*       \\
    F-score   & 54.82       & 60.82*      & 56.44        & 61.57*     \\
    Nb-copy   & 9.25        & 11.51*      & 9.94         & 12.05*\\
    \hline
    \end{tabular}
    \caption{Effectiveness analysis depending on generation size. Nb-copy: average number of words copied from the source table. 
    *: p-value $< 0.001$ for  Student T-test comparing short/long generations.}
    \label{table:parent-vs-length}
\end{table}

Considering hallucinations, where being precise should naturally be increasingly harder with sentence length, we first observe that for the pretrained model, precision  tends to decrease with longer generation (78.75 vs. 77.76), while the RL-trained model is more robust and has constant precision independently of generation length\footnote{This behavior, stopping generation early to avoid hallucinated statements, is illustrated in Figure~\ref{fig:jameslove}.}.

Regarding omissions, Table~\ref{table:parent-vs-length} shows that while both scenarios 
have similar recall/coverage scores in long generations, the RL-trained model is more relevant;
achieving a recall of $46.27$ on short generations against $44.82$ for the pretrained model.
Finally, this leads to overall higher PARENT score for the PARENTed model than the pretrained model, independently of text lengths, showing that the RL-trained model choses more accurately when to stop generating early and when to pursue longer generation, adding additional information from the table to improve coverage.

Interestingly, we find that no matter the generation size, the PARENTed model shows more reliance on the source table, as it tends to directly copy words more often than its pretrained version. 

\paragraph{Human Evaluations.}

\begin{table}
    \centering
    \footnotesize
    \begin{tabularx}{\columnwidth}{lccc}
    \toprule
    Model                  & Fluency   & Factualness & Coverage \\
    \midrule
    \texttt{Gold}                    &    91.7\%     & 32.1\%        &    4.25 \\
    \texttt{S2S+FA+RL}      &    84.5\%     & 58.3\%        & \bf4.45 \\
    \texttt{LSTM+RL}        &    \bf94\%    & \bf72.6\%     &    4.22 \\
    \bottomrule
    \end{tabularx}
    \caption{Results  of the human evaluation  on WikiBIO. The Fluency column reports the count of sentences labeled as ``fluent'' or ``mostly fluent''.}
    \label{tab:human}
\end{table}

To better measure subtleties which would not be captured by automatic metric in model outputs, we report human ratings in Table~\ref{tab:human}. Due to cost of human evaluation, we focus on the WikiBIO dataset, and three settings: the best model from the literature \texttt{S2S+FA+RL}, our model \texttt{LSTM+RL} and gold sentences. It is worth noting that our results align with \citet{dhingra2019}, as we found that around two thirds of gold references contain divergences from their associated tables.


$\bullet$ The fluency scores highlight the need for a mixed objective loss, which leverages the MLE objective ability to produce fluent output, whereas RL alone (\texttt{S2S+FA+RL}) leads to less fluent output due to the discrete metric being used as a reward. Indeed, \texttt{S2S+FA+RL} obtains only a score $84\%$, compared to gold standard of nearly $92\%$, or our model's score of $94\%$\footnote{Gold standard not being $100\%$ is explained by preprocessing choices made at dataset creation (e.g. non-english languages are not always correctly transcribed).}. 


$\bullet$  Factualness scores show that both approaches greatly improve factualness over gold standard. However, \texttt{S2S+FA+RL} still lags behind our proposed approach, which is able to leverage the PARENT-based reward to constrain the system better than would the FA module. 


$\bullet$ In contrast to factualness, \texttt{S2S+FA+RL} obtains better coverage performance than our approach, with $4.45$ vs $4.25$, showing that a component dedicated to coverage (either the FA module or the TFIDF part of the reward) leads to global outputs. Despite this, coverage is on par with gold standard.



\section{Discussion and Conclusion}
\label{sec:conclusion}

In this work, we have proposed a model-agnostic reinforcement learning framework for data-to-text aimed at reducing hallucinations and improving recall/coverage of relevant information. We shaped the reward based on PARENT \cite{dhingra2019}, which is a recently proposed metric with a high correlation with human judgement.

This 
allows for a more flexible training, where the model learns to depend less on the reference and more on the source data. Framework effectiveness is assessed via thorough experiments on two model family (RNNs and Transformers) and two benchmarks (WikiBIO and WebNLG). 
Furthermore, quantitative and qualitative evaluations show that our PARENTing framework 
obtains better results than a dedicated attention module or a less source-relying reward.


However, this approach relies on the metric employed and crafting an effective metric is still an open problem. In particular, PARENT is designed for single-entity datasets, like WikiBIO and WebNLG,
which is not reliable for more complex datasets 
containing multiple entities 
(i.g., the RotoWire dataset \citep{wiseman2017}). In this setting, the sentence ``James Harden scored 20 points.'' could achieve a high PARENT score if any player had scored $20$ points in the game. 
An interesting future work would be the design of an evaluation metric more robust to dataset peculiarities. 

\section*{Acknowledgements}
We would like to thank the H2020 project AI4EU (825619) and the ANR JCJC SESAMS (ANR-18-CE23-0001) for supporting this work.

\bibliography{main}
\bibliographystyle{acl_natbib}





\end{document}